\pdfoutput=1

\documentclass[journal]{IEEEtran}
\usepackage{graphicx}
\usepackage{subfig}
\usepackage{amsmath} 
\usepackage{epsfig}
\usepackage{breqn}
\usepackage{bm}
\usepackage{amssymb}
\usepackage{color}
\usepackage{cite}
\usepackage{balance}
\newcommand{\bb}{\bm}
\usepackage{algorithm,algorithmic}
\DeclareMathOperator*{\argminA}{arg\,min} % Jan Hlavacek
\DeclareMathOperator*{\argmaxA}{arg\,max} % Jan Hlavacek
\begin{document}
%\title{On the Assessment of Spatial Resolution of PET System with Iterative Image Reconstruction}
\title{Iterative PET Image Reconstruction Using Convolutional Neural Network Representation}

\author{Kuang Gong, Jiahui Guan, Kyungsang Kim, Xuezhu Zhang, Georges El Fakhri, Jinyi Qi* and Quanzheng Li*
\thanks{K.~Gong is with Gordon Center for Medical Imaging, Massachusetts General Hospital and Harvard Medical School, Boston, MA 02114 USA, and with the Department of Biomedical Engineering, University of California, Davis CA 95616 USA.}
\thanks{J.~Guan is with the Department of Statistics, University of California, Davis, CA 95616 USA}%
\thanks{K.~Kim, G.~Fakhri and Q.~Li* are with Gordon Center for Medical Imaging, Massachusetts General Hospital and Harvard Medical School, Boston, MA 02114 USA (email:li.quanzheng@mgh.harvard.edu)}%
\thanks{X.~Zhang and J.~Qi* are with the Department of Biomedical Engineering, University of California, Davis, CA 95616 USA (qi@ucdavis.edu)}}  
\begin{twocolumn}
\maketitle 
\begin{abstract}
PET image reconstruction is challenging due to the ill-poseness of the inverse problem and limited number of detected photons. Recently deep neural networks have been widely and successfully used in computer vision tasks and attracted growing interests in medical imaging. In this work, we trained a deep residual convolutional neural network to improve PET image quality by using the existing inter-patient information. An innovative feature of the proposed method is that we embed the neural network in the iterative reconstruction framework for image representation, rather than using it as a post-processing tool. We formulate the objective function as a constraint optimization problem and solve it using the alternating direction method of multipliers (ADMM) algorithm. Both simulation data and hybrid real data are used to evaluate the proposed method. Quantification results show that our proposed iterative neural network method can outperform the neural network denoising and conventional penalized maximum likelihood methods.
\end{abstract}


\section{Introduction}
Positron Emission Tomography (PET) is an imaging modality widely used in oncology \cite{beyer2000combined}, neurology \cite{gunn2015quantitative} and cardiology \cite{machac2005cardiac}, with the ability to observe molecular-level activities inside the tissue through the injection of specific radioactive tracers. Though PET has high sensitivity compared with other imaging modalities, its image resolution and signal to noise ratio (SNR) are still low due to various physical degradation factors and low coincident-photon counts detected. Improving PET image quality is essential, especially in applications like small lesion detection, brain imaging and longitudinal studies. Over the past decades, multiple advances have been made in PET system instrumentation, such as exploiting time of flight (TOF) information\cite{karp2008benefit}, enabling depth of interaction capability \cite{yang2009depth} and extending the solid angle coverage \cite{poon2012optimal,gong2016designing}. 

With the wide adoption of iterative reconstruction in clinical scanners, more accurate point spread function (PSF) modeling can be used to take various degradation factors into consideration \cite{gong2017sinogram}. In addition, various post processing approaches and iterative reconstruction methods have been developed by making use of local patch statistics, prior anatomical or temporal information. Denoising methods, such as the HYPR processing \cite{christian2010dynamic}, non-local mean denoising \cite{dutta2013non,chan2014postreconstruction} and guided image filtering \cite{yan2015mri} have been developed and show better performance in bias-variance tradeoff or partial volume correction than the conventional Gaussian filtering. In regularized image reconstruction,  entropy or mutual information based methods \cite{nuyts2007use,tang2009bayesian,somayajula2011pet}, segmentation based methods\cite{comtat2001clinically,baete2004anatomical}, and gradient based methods \cite{ehrhardt2016pet,knoll2016joint} have been developed by penalizing the difference between the reconstructed image and the prior information in specific domains. The Bowsher's method \cite{bowsher2004utilizing} adjusts the weight of the penalty based on similarity metrics calculated from prior images. Methods based on sparse representations \cite{chen2015sparse,tahaei2016patch,tang2016sparsity,guobao15,hutchcroft2016anatomically,novosad2016mr}, have also shown better image qualities in both static and dynamic reconstructions. Most of the aforementioned methods require prior information from the same patient which is not always available due to instrumentation limitation or long scanning time, which may hamper the practical application of these methods. Recently a new method is developed to use information in longitudinal scans \cite{ellis2017simultaneous}, but can only be applied to specific studies.

%For example, in some institutes, for some specific applications, the patient might have gone through the dynamic scanning. When combined together, the image quality of 1 hour scanning is good. For the latter patients, if they only scan like 5 minutes, the prior information should help. But currently there are no available methods used to make use of this information. 

In this paper, we explore the potential of using existing inter-patient information via deep neural network to improve PET image reconstruction. Over the past several years, deep neural networks have been widely and successfully applied to computer vision tasks, such as image segmentation \cite{ronneberger2015u}, object detection \cite{ren2015faster} and image super resolution \cite{dong2016image}, due to the availability of large data sets, advances in optimization algorithms and emerging of effective network structures. Recently, it has been applied to medical imaging, such as image denoising and artifact reduction, using convolutional neural network (CNN) \cite{wang2016accelerating,kang2016deep,chen2017low,wu2017cascaded} or generative adversarial network (GAN) \cite{wolterink2017generative}. It showed comparable or superior results to the iterative reconstruction but at a faster speed. In this paper, we propose a new framework to integrate deep CNN in PET image reconstruction. The network structure is a combination of U-net structure \cite{ronneberger2015u} and the residual network \cite{he2016deep}. Different from existing CNN based image denoising methods, we use a CNN trained with iterative reconstructions of low-counts data as the input and high-counts reconstructions as the label to represent the unknown PET image to be reconstructed. Rather than feeding a noisy image into the CNN, we use the CNN to define the feasible set of valid PET images. To our knowledge, this is the first of its kind in the applications of neural network in medical imaging. The solution is formulated as the solution of a constrained optimization problem and sought by using the alternating direction method of multipliers (ADMM) algorithm \cite{boyd2011distributed}. The proposed method is validated using both simulation and hybrid real data.

%
%The implemented network structure is a combination of U-net structure \cite{ronneberger2015u} and the residual network \cite{he2016deep}. Different from CT by using FBP reconstruction as the input, here we use the iterative reconstruction based on low dose as the input and high dose reconstruction as the label in the training process. Besides, to overcome the over-smooth phenomenon caused by applying CNN, we propose to add a data consistence constraint to further improve the results by incorporating the neural network into the iterative reconstruction framework. The alternating direction method of multipliers (ADMM) algorithm \cite{boyd2011distributed} is used to solve the optimization problem. Based on quantification results using both simulation and  hybrid real data sets, the proposed method can outperform the CNN denoising and other commonly used methods. 

%One challenge in deep learning is the requirement of a large number of training images. In CT or MRI applications, high dose or fully sampled images are often available as the target image and low dose or partially sampled images are used as the input. In PET, such high-dose images are not readily available. In our work, we used PET images reconstructed from one-hour scan as the target image and the images reconstructed from downsampled low-count data as the input.

The main contributions of this paper include (1) using dynamic data of prior patients to train a network for PET denoising and (2) proposing to incorporate the neural network into the iterative reconstruction framework and demonstrating better performance than the denoising approach. This paper is organized as follows. Section 2 introduces the theory and optimization algorithm. Section 3 describes the Monte Carlo simulations and real data used in the evaluation. Experimental results are shown in Section 4, followed by discussions in Section 5. Finally conclusions are drawn in Section 6.

\section{Theory}
\subsection{PET data model}
In PET image reconstruction, the measured data $\bb{y} \in \mathbb{R}^{M \times 1} $ can be modeled as a collection of independent Poisson random variables and its mean  $\bar{\bb{y}} \in \mathbb{R}^{M \times 1} $ is related to the unknown image $\bb{x} \in \mathbb{R}^{N \times 1}$ through an affine transform
\begin{equation}
\bar{\bb{y}} = \bb{P}\bb{x} + \bb{s} + \bb{r},
\label{eqn_mean}
\end{equation}
where $\bb{P} \in \mathbb{R}^{M \times N}$ is the detection probability matrix, with $P_{ij}$ denoting the probability of photons originating from voxel $j$ being detected by detector $i$ \cite{qi1998high}. $\bb{s} \in \mathbb{R}^{M \times 1}$ denotes the expectation of scattered events, and $\bb{r} \in \mathbb{R}^{M \times 1}$ denotes the expectation of random coincidences. $M$ is the number of lines of response (LOR) and $N$ is the number of pixels in image space.  The log-likelihood function can be written as
\begin{equation}
L(\bb{y}|\bb{x}) = \sum_{i=1}^M y_i \log \bar{y}_i - \bar{y}_i - \log y_i!.
\label{likelihood}
\end{equation}
The maximum likelihood estimate of the unknown image $\bb{x}$ can be found by
\begin{equation}
\hat{\bb{x}} = \mbox{arg} \max_{\bb{x} \ge 0} L(\bb{y}|\bb{x}).
\end{equation}
%\subsection{Convolutional neural network}
%For convolutional neural network, the $i$th output feature of  each convolution layer,  ${\bf x}^{\mbox{conv}}_i  \in \mathbb{R}^{L \times L }$, can be represented by 
%\begin{equation}
%{\bf x}^{\mbox{conv}}_i = \bb{w}_i \ast {\bf x}^{\mbox{in}} + b_i,
%\end{equation}
%where ${{\bf x}^{\mbox{in}} \in \mathbb{R}^{L \times L \times c}}$ is the input with spatial size $L \times L$ and channel size $c$, $\bb{w} \in \mathbb{R}^{l \times l \times c}$ is the spatial invariant filter with spatial size $l$ and depth size $c$, $b_i$ is the spatial independent bias item. The output of the convolution layer is sent through the nonlinear activation function. Here we use the Rectified Linear Unit (ReLU) and the output is 
%\begin{equation}
%{\bf x}^{\mbox{relu}}_i = \max(0,{\bf x}^{\mbox{conv}}_i).
%\end{equation}
%\begin{figure}[t]
%\centering
%\subfloat{\includegraphics[trim=4.3cm 0cm 4cm 0cm, clip, width=6in]{unet_structure_stride2V5.pdf}}
%\caption{The schematic diagram of the neural network architecture.}
%\label{fig:u-net-structure}
%\end{figure}
%If the image size in the following layer needs to change, a convolution layer with a specific stride size is used to down-sample the input and the transposed convolution layer is used to up-sample the input. Besides, in order to stablize the optimization and overcome the gradient vanishing problem, a batch normalization layer is always added ahead of the ReLu layer \cite{ioffe2015batch}. These are basic layers to construct a CNN.

\subsection{Representing PET images using neural network}
Previously, the kernel method \cite{guobao15} used a kernel representation $\bb{x} = K\bb{\alpha}$ to represent the image $\bb{x}$, through which the prior temporal or anatomical information can be embedded into the kernel matrix $\bb{K}  \in \mathbb{R}^{N \times N }$. Inspired by this idea, here we represent the unknown image $\bb{x}$ as
\begin{equation}
 \bb{x} = f(\bb{\alpha}), 
\label{eq:sparse-represent}
\end{equation}
where $f:\mathbb{R}\rightarrow\mathbb{R}$ represents the neural network and $\bb{\alpha}$ denotes the input to the neural network. Through this representation, inter-patient information and intra-patient information can be included into the iterative reconstruction framework through pre-training the neural network using existing data.

\begin{figure*}[t]
\centering
\subfloat{\includegraphics[trim=0cm 3cm 0cm 0cm, clip, width=7.2in]{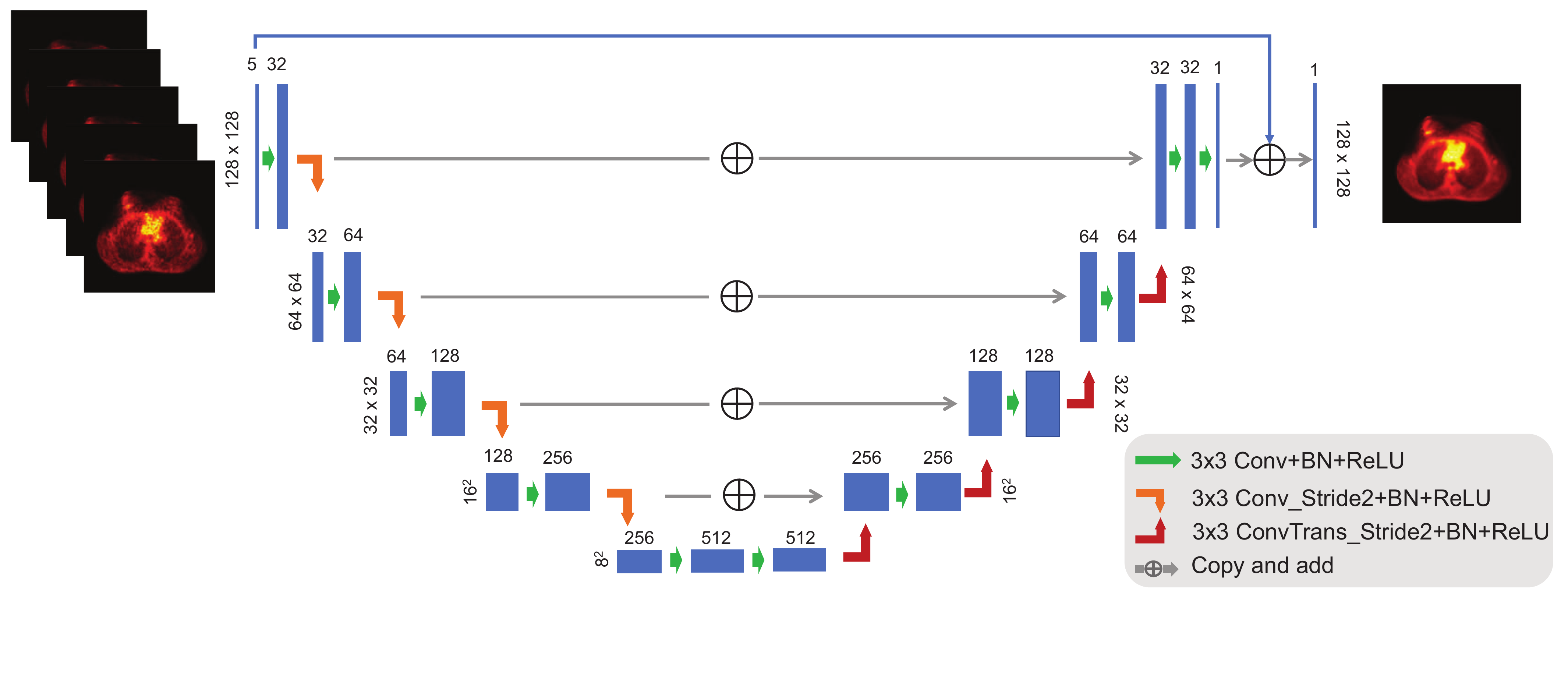}}
\caption{The schematic diagram of the neural network architecture.}
\label{fig:u-net-structure}
\end{figure*}

Our network implemented in this work is based on the U-net structure  \cite{ronneberger2015u} and also includes the batch normalization layer. The overall network architecture is summarized in Fig.~\ref{fig:u-net-structure}. It consists of repetitive applications of 1) $3$x$3$ convolutional layer, 2) batch normalization layer, 3) ReLU layer, 4) convolutional layer with stride 2 for down-sampling, 5) transposed convolutional layer with stride 2 for up-sampling, and 6) identity mapping layer that adds the left-side feature layer to the right-side. In our implementation, there are three major modifications compared to the original U-net:
\begin{enumerate}[]
\item using convolutional layer with stride 2 to down-sample the image instead of using max pooling layer, to construct a fully convolutional network;
\item directly adding the left side feature to the right side instead of concatenating, to reduce the number of training parameters;
\item connecting the input to the output, to construct a residual network \cite{he2016deep}. 
\end{enumerate}
The left-hand side of the architecture aims to compress the input path layer by layer, an ``encoder" part, while the right-hand side is to expand the path, a ``decode'' part.  This neural network has 19 convolutional layers in total and the largest feature size is 512. To reduce computational cost, the network denoises PET image one slice at a time. However, the input layer has five channels to provide information from 4 neighboring axial slices for effective noise removal. We have found that if the input did not contain the neighboring axial information, there will be artifacts in the axial direction. The network is trained with reconstructed images from low counts data as the input and the images reconstructed from high counts data as the label.

When substituting the representation in (\ref{eq:sparse-represent}) using the above mentioned network structure, the original PET system model shown in (\ref{eqn_mean}) can be rewritten as  
\begin{equation}
\bar{\bb{y}} = \bb{P}f(\bb{\alpha}) + \bb{s} + \bb{r}.
\label{eqn_mean_alpha}
\end{equation}
The maximum likelihood estimate of the unknown image $\bb{x}$ can be calculated as
\begin{align}
\hat{\bb{x}} &= f(\hat{\bb{\alpha}}), \\
\hat{\bb{\alpha}} &= \mbox{arg} \max_{\bb{\alpha} \ge 0} L(\bb{y}|\bb{\alpha}). \label{eq:max_alpha}
\end{align}
The objective function in (\ref{eq:max_alpha}) is difficult to solve due to the nonlinearity of the neural network representation. Here we transfer it to the constrained format as below

\begin{equation}
\begin{aligned}
& {\text{max}}
& & L(\bb{y}|{\bb{x}})  \\
& \text{s.t.}
& & {\bb{x}} = f(\bb{\alpha}). 
\end{aligned}
\label{eq:contrained_optim}
\end{equation}

\subsection{Optimization}
We use the Augmented Lagrangian format for the constrained optimization 
problem in (\ref{eq:contrained_optim}) as
\begin{equation}
L_{\rho} = L(\bb{y}|{\bb{x}}) - \frac{\rho}{2} \Vert{\bb{x}} -{f}({\bb{\alpha}})+ \bb{\mu}\Vert^2 + \frac{\rho}{2} \Vert \bb{\mu} \Vert^2,
\label{admm_scale}
\end{equation}
which can be solved by the ADMM algorithm iteratively in three steps
\begin{align}
\bb{x}^{n+1} &= \argmaxA_{\bb{x}}  L(\bb{y}|\bb{x}) - \frac{\rho}{2} \Vert\bb{x} - {f}({\bb{\alpha}^n}) + \bb{\mu}^n\Vert^2, \label{eq:sub1}\\
\bb{\alpha}^{n+1} &= \argminA_{\bb{\alpha}} \Vert {f}({\bb{\alpha}})- (\bb{x}^{n+1} + \bb{\mu}^n)\Vert^2,  \label{eq:sub2} \\
\bb{\mu}^{n+1} &= \bb{\mu}^n + \bb{x}^{n+1} - f({\bb{\alpha}^{n+1}}).  \label{eq:sub3}
\end{align} 
Subproblem (\ref{eq:sub1}) is a penalized PET reconstruction problem. We solve it using the optimization transfer method\cite{wang2012penalized}. As $\bb{x}$ in $L(\bb{y}|\bb{x})$ is coupled together, we first construct a surrogate function $Q_{L}(\bb{x}|\bb{x}^n)$ for $L(\bb{y}|\bb{x})$ as follows
\begin{equation}
Q_{L}(\bb{x}|\bb{x}^n) =  \sum_{j=1}^{n_j}p_j(\hat{x}^{n+1}_{j,\text{EM}}\log x_j - x_j),
\label{admm_scale}
\end{equation}
where $p_j = \sum_{i=1}^{n_i}p_{ij}$ and $\hat{x}^{n+1}_{j,\text{EM}}$ is calculated by 
\begin{equation}
\hat{x}^{n+1}_{j,\text{EM}} = \frac{x^n_j}{p_j} \sum_{i=1}^{n_i}p_{ij}\frac{y_i}{ [\bb{P}\bb{x}^n]_i + s_i + r_i}.
\label{admm_scale}
\end{equation}
It can be verified that the constructed surrogate function $Q_{L}(\bb{x}|\bb{x}^n)$ fulfills the following two conditions:
\begin{align}
Q_{L}(\bb{x}; \bb{x}^n)-Q_{L}(\bb{x}^n; \bb{x}^n) &\leq  L(\bb{y};\bb{x})-L(\bb{y}; \bb{x}^n), \\
\nabla Q_{L}(\bb{x}^n; \bb{x}^n) &= \nabla L(\bb{y}; \bb{x}^n).
\label{opt_transfer_cond}
\end{align}
After getting this surrogate function, subproblem (\ref{eq:sub1}) can be optimized pixel by pixel. For pixel $j$, the surrogate objective function for subproblem (\ref{eq:sub1}) is 
\begin{equation}
P(x_j|\bb{x}^n) = p_j(\hat{x}^{n+1}_{j,\text{EM}}\log x_j - x_j)  - \frac{\rho}{2}\left[x_j - f(\bb{\alpha})^n_j + \bb{\mu}_j^n\right]^2. 
\label{eq:surrogate-pixel}
\end{equation}
The final update equation for pixel $j$ after maximizing (\ref{eq:surrogate-pixel}) is 
\begin{align}
\hat{x}^{n+1}_{j}  &= \frac{1}{2}\bigg[f(\bb{\alpha}^n)_j - \mu^n_j -p_j / \rho \nonumber \nonumber \\
&+ \sqrt{(f(\bb{\alpha}^n)_j - \mu^n_j - p_j/ \rho)^2 + 4\hat{x}^{n+1}_{j,\text{EM}} p_j /\rho)^2} \bigg].
\label{update_sub1}
\end{align}
Subproblem (\ref{eq:sub2}) is a non-linear least square problem. In order to solve it, we need to compute the gradient of the objective function with respect to the input $\bb{\alpha}$. As it is difficult to calculate the Jacobian matrix or Hessian matrix of the objective function with respect to the input in current network platform, we use a first-order method as follows
\begin{equation}
{\alpha}^{n+1}_{j} = {\alpha}^{n}_{j} - L \frac{\partial{f(\bb{\alpha}^n)_j}}{\partial{\alpha_j}}[f(\bb{\alpha}^n)_j - x_j^{n+1} - u_j^n], 
\label{eq:update-sub2}
\end{equation}
where $L$ is the step size. In our implementation, $L$ was chosen so that the objective function in subproblem (\ref{eq:sub2}) can be monotonic decreasing. For our neural network, the input have five channels, which include four neighboring axial slices. Therefore, equation (\ref{eq:update-sub2}) should be modified to include the first-order gradients from the other four neighboring slices. The final update equation is changed to
\begin{equation}
{\alpha}^{n+1}_{j} = {\alpha}^{n}_{j} - L \sum_{c=1}^{N_c}\frac{\partial{f(\bb{\alpha}^n)_{j}}}{\partial{\alpha_{j-m}}}[f(\bb{\alpha}^n)_{j-m} - x_{j-m}^{n+1} - u_{j-m}^n], 
\label{admm_scale_final}
\end{equation}
where $N_c=5$ is the number of channels, $m = [c - (N_c-1)/2]n_{t}^2$, and $n_{t}=128^2$ is the spatial input size. In order to accelerate the convergence speed, Nesterov momentum method was used in subproblem (\ref{eq:sub2}) \cite{nesterov1983method}. In our implementation, we run one iteration for subproblem (\ref{eq:sub1}) and then run five iterations for subproblem (\ref{eq:sub2}). As subproblem (\ref{eq:sub2}) is a non-linear problem, it is very easy to be trapped into a local minimum and it is thus essential to assign a good initial for $\bb{\alpha}$. In our implementation, we first ran MLEM for 30 iterations and used its CNN output as the initial for $\bb{\alpha}$. The overall algorithm flowchart is presented in Algorithm 1.  
 \begin{algorithm}[t]
 \caption{{Algorithm for iterative PET reconstruction incorporating convolutional neural network}}
 \begin{algorithmic}[1]
 \renewcommand{\algorithmicrequire}{\textbf{Input:}}
 \renewcommand{\algorithmicensure}{abf{Output:}}
 \REQUIRE Maximum iteration number \texttt{MaxIt}, 
 		sub-iteration number \texttt{SubIt}\\
  \STATE Initialize $\bb{\alpha}^{0,\texttt{SubIt}} = f(\bb{x}_{\mbox{\tiny EM}}^{30})$
  \FOR {$n = 1$ to \texttt{MaxIt}}
  \STATE $\hat{x}^{n}_{j,\mbox{\tiny EM}} = \frac{x^{n-1}_j}{p_j} \sum_{i=1}^{n_i}p_{ij}\frac{y_i}{ [\bb{P}\bb{x}^{n-1}]_i + s_i + r_i}$, where $p_j = \sum_{i=1}^{n_i}p_{ij}$
  \STATE $\hat{x}^{n}_{j} = \frac{1}{2}\Bigl[f(\bb{\alpha}^{n-1})_j - \mu^{n-1}_j -p_j / \rho$ \\ $+ \sqrt{(f(\bb{\alpha}^{n-1})_j - \mu^n_j - p_j/ \rho)^2 + 4\hat{x}^{n}_{j,\mbox{\tiny EM}} p_j / \rho)^2}\biggr]$
  \STATE $\bb{\alpha}^{n,0} = \bb{\alpha}^{n-1,\texttt{SubIt}}$, $\bb{\theta}^{n,0} = \bb{\alpha}^{n-1,\texttt{SubIt}}$,$t_0 = 1$
	\FOR {$k = 1$ to \texttt{SubIt}}
		\STATE $t_{k} = (1+\sqrt{\smash 1+4t^2_{k-1}})/2$
		\STATE ${\alpha}^{n,k}_{j} = {\alpha}^{n,k-1}_{j} -$\\ $L \sum_{c=1}^{N_c}\frac{\partial{f(\bb{\theta}^{n,k-1})_{j}}}{\partial{\theta^{n,k-1}_{j}}}[f(\bb{\theta}^{n,k-1})_{j-m} - x_{j-m}^{n} - u_{j-m}^n]$
		\STATE ${\bb{\theta}}^{n+1} = {\bb{\alpha}}^{n+1} + \frac{t_n-1}{t_{n+1}}({\bb{\alpha}}^{n+1} - {\bb{\alpha}}^{n}) $
		
 	\ENDFOR
 \ENDFOR
 \RETURN $\hat{\bb{x}} =f(\hat{\bb{\alpha}}^{\texttt{MaxIt}, \texttt{SubIt}})$ 
 \end{algorithmic} 
 \end{algorithm}
 
\subsection{Implementation details and reference methods}
The neural network was implemented using TensorFlow 1.0 on a NVIDIA GTX 1080Ti. The network input size is $128 \times 128 \times 5$ and the output size is $128 \times 128$. During training, Adam algorithm \cite{kingma2014adam} was used as the optimizer and the cost function was calculated as the L2 norm between the network outputs and the label images. The first-order gradient used in subproblem (\ref{eq:sub2}) was implemented using the tf.gradient function in TensorFlow.

We compared the proposed methods with the post-reconstruction Gaussian filtering and a penalized reconstruction. For the penalized reconstruction, the fair penalty was used with the form 
\begin{equation}
\phi_{\mbox{fair}}(t) = \sigma \left[ \frac{|t|}{\sigma} - \log \left(1+\frac{|t|}{\sigma}\right)\right].
\label{eq:fair}
\end{equation} 
The fair penalty approaches the L-1 penalty when $\sigma  \ll  |t|$ and is similar to the quadratic penalty when $\sigma \gg |t|$. In our implementation, $\sigma$ was set to be $1\mathrm{e}{-5}$ of the mean image intensity in order to have the edge preserving capability. MAP EM algorithm was used in the penalized reconstruction \cite{wang2012penalized}. In order to accelerate the convergence, 10 iterations of MLEM algorithm was used for ``warming up'' before running the MAP EM algorithm.

\section{Experimental setup}
\subsection{Simulation study}

The computer simulation modeled the geometry of a GE 690 scanner \cite{bettinardi2011physical}.
The scanner consists of $13,824$ LYSO crystals forming a ring of diameter of 81 cm with an axial field of view (FOV) of 157 mm. The crystal size for this scanner is $4.2 \times 6.3 \times 25$ $\mbox{mm}^3$. Nineteen XCAT phantoms with different organ sizes and genders were employed in the simulation \cite{segars20104d}. Apart from the major organs, thirty hot spheres of diameters ranging from 12.8 mm to 22.4 mm were inserted into eighteen phantoms as lung lesions   for the training images. For the test image, five lesions with diameter 12.8 mm were inserted. The time activity curves (TAC) of different tissues were generated mimicking an FDG scan using a two-tissue-compartment model with an analytic blood input function \cite{feng1997technique}. In order to simulate the population difference, each kinetic parameter was modelled as a Gaussian variable with coefficient of variation equal to 0.1. The mean values of the kinetic parameters are presented in Table~\ref{tab:kinetics} \cite{qiao2011dynamic,karakatsanis2013dynamic}. The TACs using the mean kinetic parameters are shown in Fig.~\ref{fig:simu-tac}. The system matrix $\bb{P}$ used in the data generation and image reconstruction was computed by using the multi-ray tracing method \cite{zhou2011fast}, which modeled the inter-crystal photon penetration. The image matrix size is 128 $\times$ 128 $\times$ 49 and the voxel size is 3.27 $\times$ 3.27 $\times$ 3.27 $\mbox{mm}^3$. Noise-free sinogram data were generated by forward-projecting the ground-truth images using the system matrix and the attenuation map. Poisson noise was then introduced to the noise-free data by setting the total count level to be equivalent to an 1-hour FDG scan with 5 mCi injection. Uniform random and scatter events were simulated and accounted for 60\% of the noise free prompt data in all time frames to match those observed in real data-sets. During image reconstruction, all the correction factors were assumed to be known exactly. 
\begin{table}[t]
\caption{The mean values of the simulated kinetic parameters of FDG for different organs. $V$ stands for blood volume ratio.}
\label{tab:kinetics}
\centering
\begin{tabular}{c|c|c|c|c|c}
    \hline \hline 
 	 Tissue & $K_1$ & $k_2$ & $k_3$ & $k_4$ &  $V$ \\
    \hline
   	Myocardium &  $0.6$ & $1.2$ & $0.1$ & $0.001$ & $0$ \\
	\hline
    Liver &  $0.864$ & $0.981$ & $0.005$ & $0.016$ & $0$   \\
	\hline
	Lung &  $0.108$ & $0.735$ & $0.016$ & $0.013$ & $0.017$   \\
	\hline
	Kidney &  $0.263$ & $0.299$ & $0$ & $0$ & $0.438$   \\
	\hline
	Spleen &  $1.207$ & $1.909$ & $0.008$ & $0.014$ & $0$   \\
	\hline
	Pancreas &  $0.648$ & $1.64$ & $0.027$ & $0.016$ & $0.107$   \\	
	\hline
	Muscle/Bone/Soft tissue &  $0.047$ & $0.325$ & $0.084$ & $0$ & $0.019$   \\
	\hline
	Marrow &  $0.425$ & $1.055$ & $0.023$ & $0.013$ & $0.04$   \\
	\hline
	Lung lesion &  $0.63$ & $0.842$ & $0.092$ & $0.014$ & $0.132$   \\
    \hline \hline
\end{tabular}
\end{table}

\begin{figure}[t]
\centering
\subfloat{\includegraphics[trim=1.2cm 0.3cm 1cm 1cm, clip, width=3.3in]{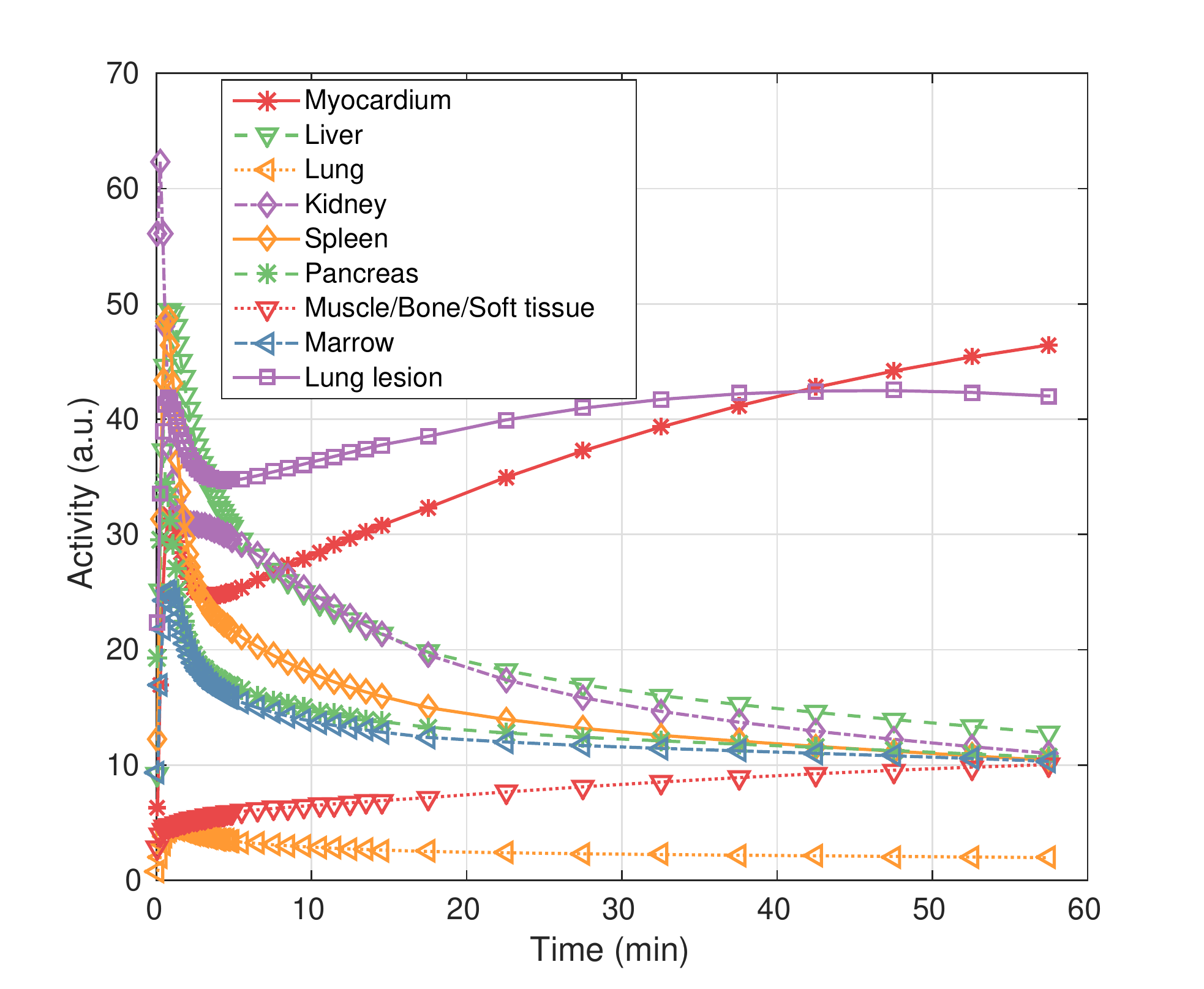}}
\caption{The simulated time activity curves based on the kinetic parameters shown in Table~\ref{tab:kinetics}.}
\label{fig:simu-tac}
\end{figure}

To generate the training data, forty-minutes data from 20 min to 60 min post injection were combined into a high count sinogram  and reconstructed as the label image for training. The high count data was down-sampled to 1/10th of the counts and reconstructed as the input image. In order to account for different noise levels, images reconstructed at iteration 20, 40, 60 using ML EM algorithm were all used in the training phase. In total $49$ ($\#$ of slices per phantom) $\times$ $18$ ($\#$ of phantoms) $\times$ $3$ ($\#$ of different iterations) training data pairs were generated. Different rotations and translations were applied to each training pair to enable larger data capacity for the training.  The training data set was separated randomly into 45 batches for every epoch. In total 1000 epochs were run. Three training pair examples are shown in Fig.~\ref{fig:simu-train-pair}.

During the evaluation, 20 low-counts realizations of the testing phantom, generated by pooling the last 40 min data together and resampling with a 1/10 ratio,  were reconstructed using different methods. For quantitative comparison, contrast recovery (CR) vs.\ the standard deviation (STD) curves were plotted. The CR was computed from the lung lesion regions as 
\begin{equation}
\mbox{CR} = \frac{1}{R}\sum_{r = 1}^{R}\bar{a}_{r}/{a}^{\mbox{true}},
\label{eq:crc}
\end{equation}
where $R=20$ is the number of realizations, $\bar{a}_r$ is the average uptake of all the lung lesions in the test phantom. The background STD was computed as
\begin{equation}
\mbox{STD} = \frac{1}{K_b}\sum_{k = 1}^{K_b}\frac{\sqrt{\frac{1}{R-1}\sum_{r=1}^R(b_{r,k} - \bar{b}_k)^2}}{\bar{b}_k}, 
\label{eq:std}
\end{equation}
where $\bar{b}_k = 1/{R}\sum_{r=1}^R b_{r,k}$ is the average of the background ROI means over realizations, and $K_b=42$ is the total number of background ROIs chosen.  

\begin{figure}[t]
\centering
\subfloat{\includegraphics[trim=2.4cm 0.6cm 3cm 0.9cm, clip, width=3.7in]{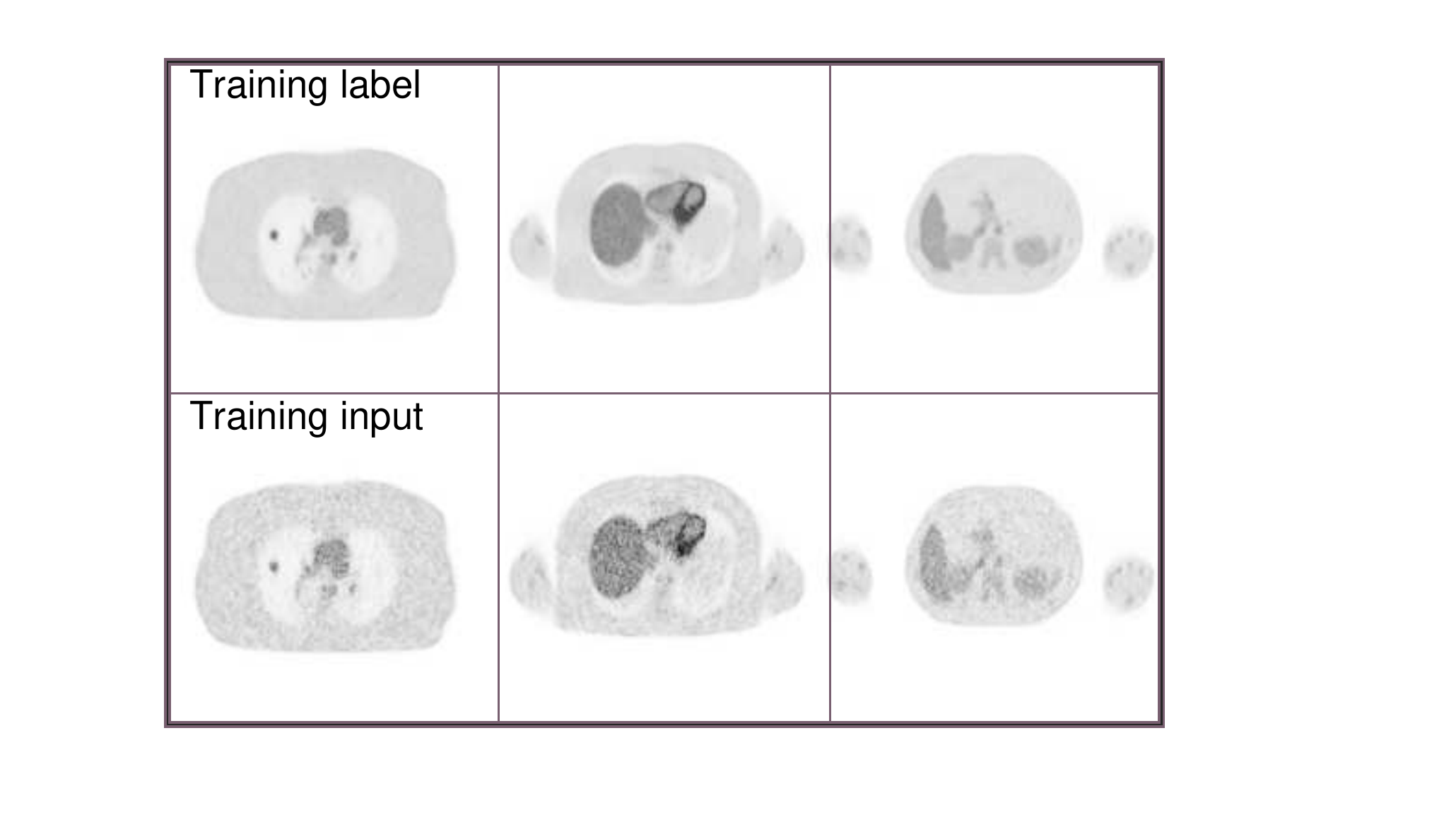}}
\caption{Three example slices of the training data pairs from three simulated XCAT phantoms.}
\label{fig:simu-train-pair}
\end{figure}

\subsection{Hybrid real data}
Six patient data sets of one hour FDG dynamic scan acquired on a GE 690 scanner with 5 mCi dose injection were employed in this study. Training and validation data were generated in the same way as that in the simulation. The system matrix used in the reconstruction is the same as the one used in the simulation. Normalization, attenuation correction, randoms and scatters were generated using the manufacturer software and included in image reconstruction. Five patient data sets were used in the training and the last one was left for validation. As no ground truth exist in the real data-sets, 27 lesions were inserted in the training data and 5 in the testing data to generate the hybrid real data-sets for quantitative analysis. The diameters of the lesions inserted into the training data sets range from 12.8 mm to 22.4 mm and the diameter for the lesions inserted in the testing data is 12.8 mm. The intensity of all the lesions were simulated as a Gaussian random variable with coefficient of variation equal to 0.2 to simulate the population difference.  In order to increase the training samples, for each patient data set we have generated five low-dose realizations from the high-counts data. Training pairs of iteration 20, 40, 60 were also included to account for different noise levels. In total $49$ ($\#$ of slices per data set) $\times$ $5$ ($\#$ of patients) $\times$ $3$ ($\#$ of different iterations)$\times$ $5$ ($\#$ of realizations) training data sets were generated with different rotations and translations. Three pairs of the training examples are shown in Fig.~\ref{fig:real-train-pair}.

Twenty realizations of the low dose data sets were resampled from the testing data  and reconstructed to evaluate the noise performance. Forty-seven background ROIs were chosen in the liver region to calculate the STD as presented in (\ref{eq:std}).  For lesion quantification, images with and without the
inserted lesion were reconstructed and the difference was taken
to obtain the lesion only image and compared with the ground
truth. The lesion contrast recovery was calculated as in (\ref{eq:crc}).

\begin{figure}[t]
\centering
\subfloat{\includegraphics[trim=2.4cm 0.6cm 3cm 0.9cm, clip, width=3.7in]{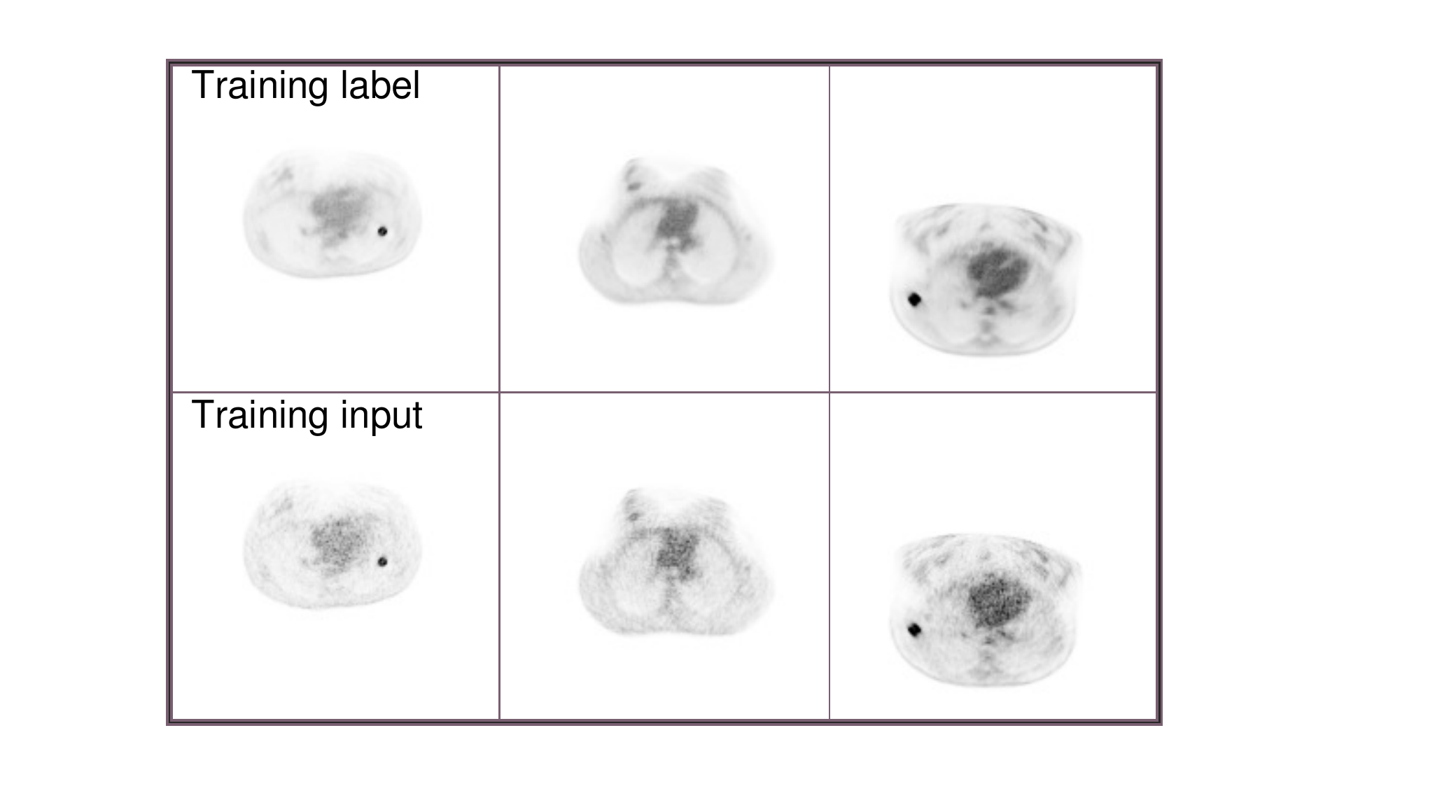}}
\caption{Three example slices of the training data pairs from three real patient scanning.}
\label{fig:real-train-pair}
\end{figure}

%\begin{figure*}[h]
%\centering
%\subfloat[]{\includegraphics[trim=0cm 0cm 0cm 0cm, clip, width=3.3in]{compare_connection-eps-converted-to.pdf}}
%\subfloat[]{\includegraphics[trim=0cm 0cm 0cm 0cm, clip, width=3.3in]{compare_plain-eps-converted-to.pdf}} \\
%\caption{(a) Comparison between using U-net and the U-net without the identify adding connection; (b) Comparison between U-net and the plain CNN with 16 features and 32 features. }
%\label{fig:network-compare}
%\end{figure*}

\begin{figure*}[h]
\centering
\subfloat{\includegraphics[trim=4.8cm 9.8cm 4.8cm 9cm, clip, width=8in]{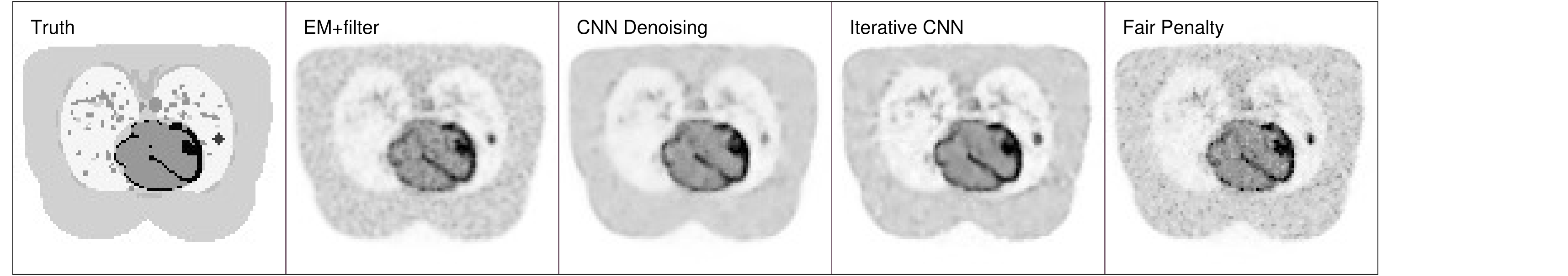}} \vspace{-0.3cm}\\
\subfloat{\includegraphics[trim=4.8cm 11.2cm 4.8cm 10.5cm, clip, width=8in]{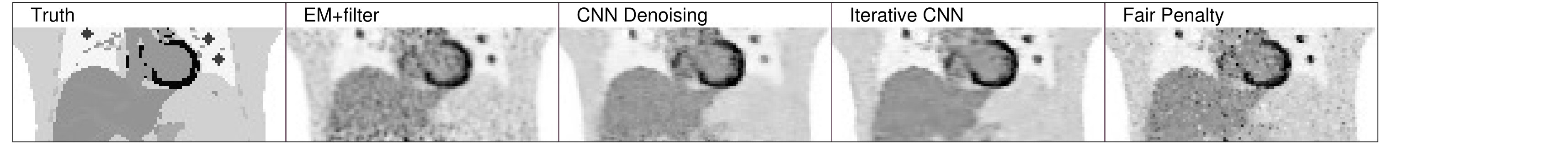}} \vspace{-0.3cm}\\
\subfloat{\includegraphics[trim=4.8cm 11.2cm 4.8cm 10.5cm, clip, width=8in]{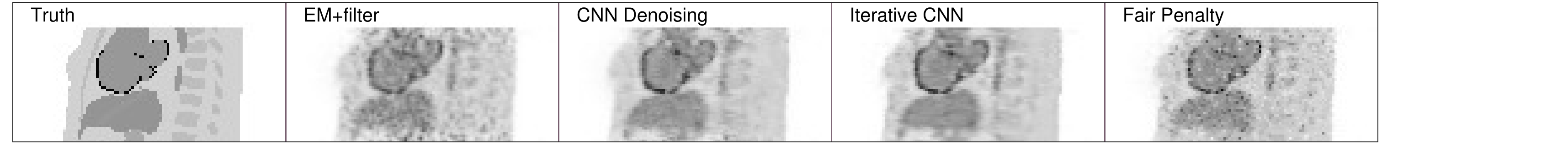}} \\
\caption{Three views of the reconstructed images using different methods for the simulation data set. From left to right: Ground truth, Gaussian denoising, CNN denoising, iterative CNN  reconstruction and Penalized reconstruction. }
\label{fig:simu-image}
\end{figure*}
\section{Results}

\subsection{Simulation results}
Fig.~\ref{fig:simu-image} shows three orthogonal slices of the reconstructed images using different methods. From the image appearance, we can see that the proposed iterative CNN method can generate images with a higher lung lesion uptake and reveal more vessel details in the lung region as compared with the CNN denoising method. This is beneficial as the CNN method is criticized for over-smoothing and losing small structures due to the L2 norm used as the cost function. Both CNN approaches are better than the traditional Gaussian post filtering method as the images have less noise but also keep all the detailed features, such as the thin myocardium regions. The penalized reconstruction result has a high lesion uptake, but also has some noise spots showing up in different regions. These observations are consistent with the quantitative results shown in Fig.~\ref{fig:simu-crc-std}. In terms of the bias-variance trade-off, the proposed iterative CNN method has the best performance among all methods.

\begin{figure}[t]
\centering
\subfloat{\includegraphics[trim=0cm 0cm 0cm 0cm, clip, width=3.3in]{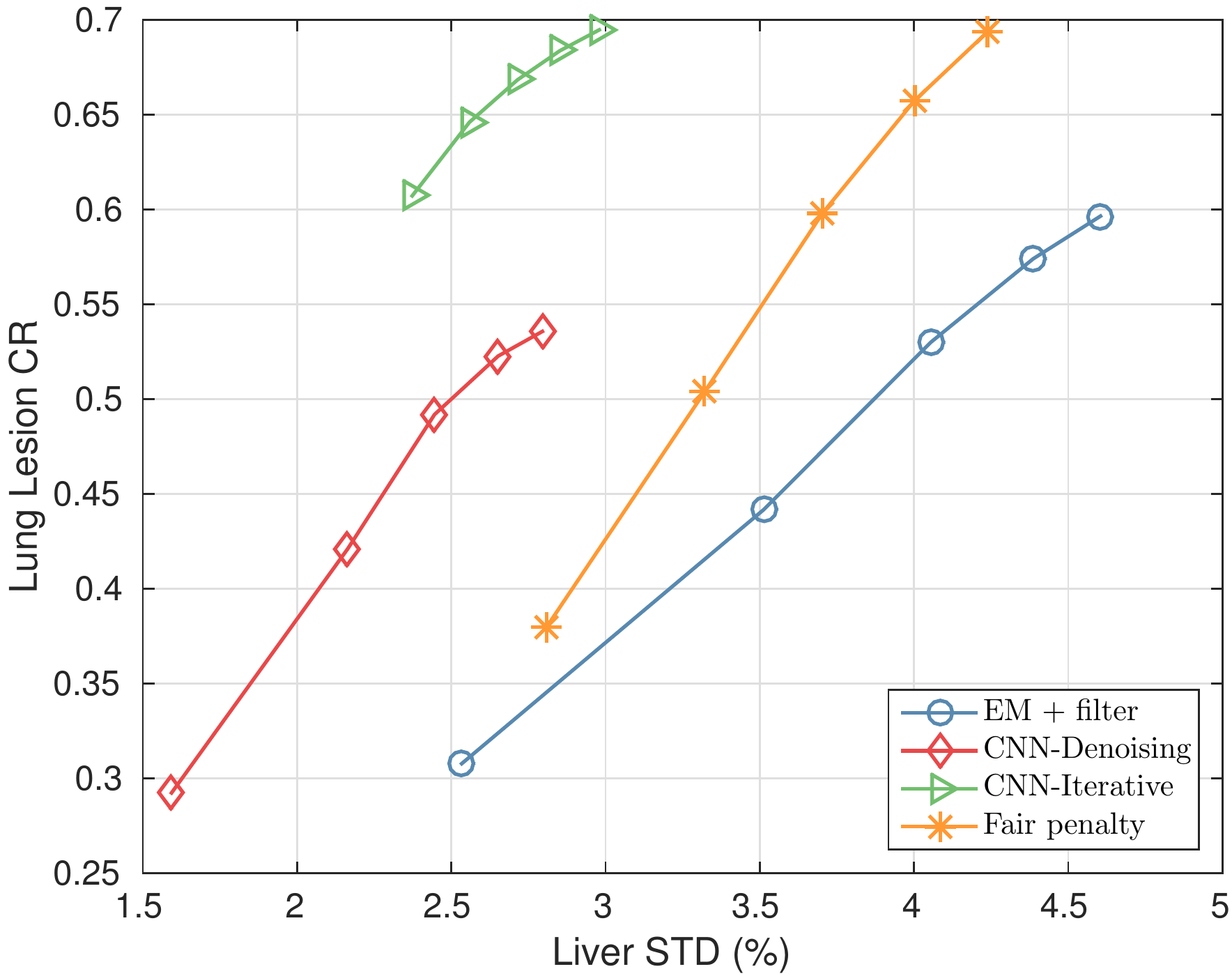}}
\caption{The contrast recovery vs. STD curves using different methods for the simulated data sets. }
\label{fig:simu-crc-std}
\end{figure}
\subsection{Real data results}
\begin{figure*}[htp]
\centering
\subfloat{\includegraphics[trim=4.8cm 9cm 4.8cm 8cm, clip, width=8in]{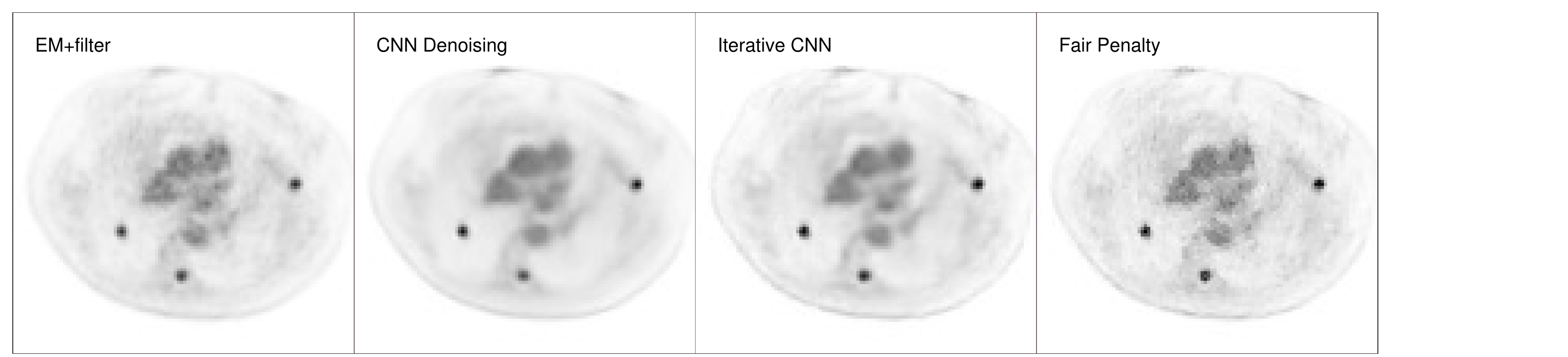}}\vspace{-0.2cm} \\
\subfloat{\includegraphics[trim=4.8cm 10.8cm 4.8cm 10.1cm, clip, width=8in]{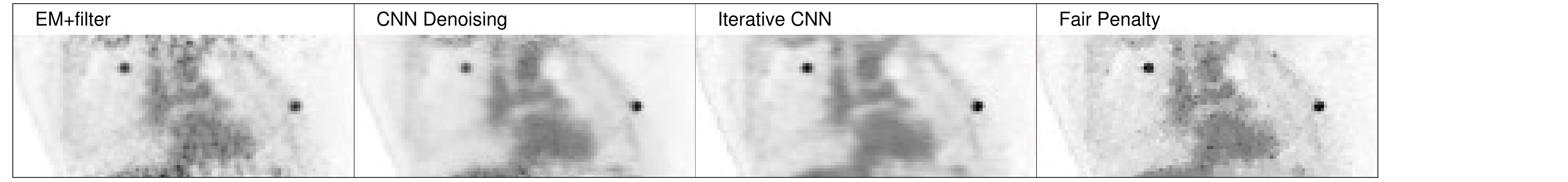}} \vspace{-0.2cm}\\
\subfloat{\includegraphics[trim=4.8cm 10.8cm 4.8cm 10.1cm, clip, width=8in]{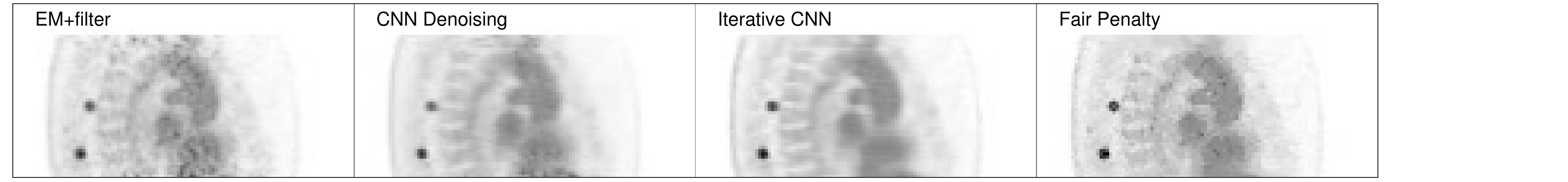}} \\
\caption{Three views of the reconstructed images using different methods for the hybrid real data set. From left to right: Ground truth, Gaussian denoising, CNN denoising, iterative CNN  reconstruction and penalized reconstruction.}
\label{fig:real-image}
\end{figure*}

\begin{figure}[!t]
\centering
\subfloat{\includegraphics[trim=0cm 0cm 0cm 0cm, clip, width=3.3in]{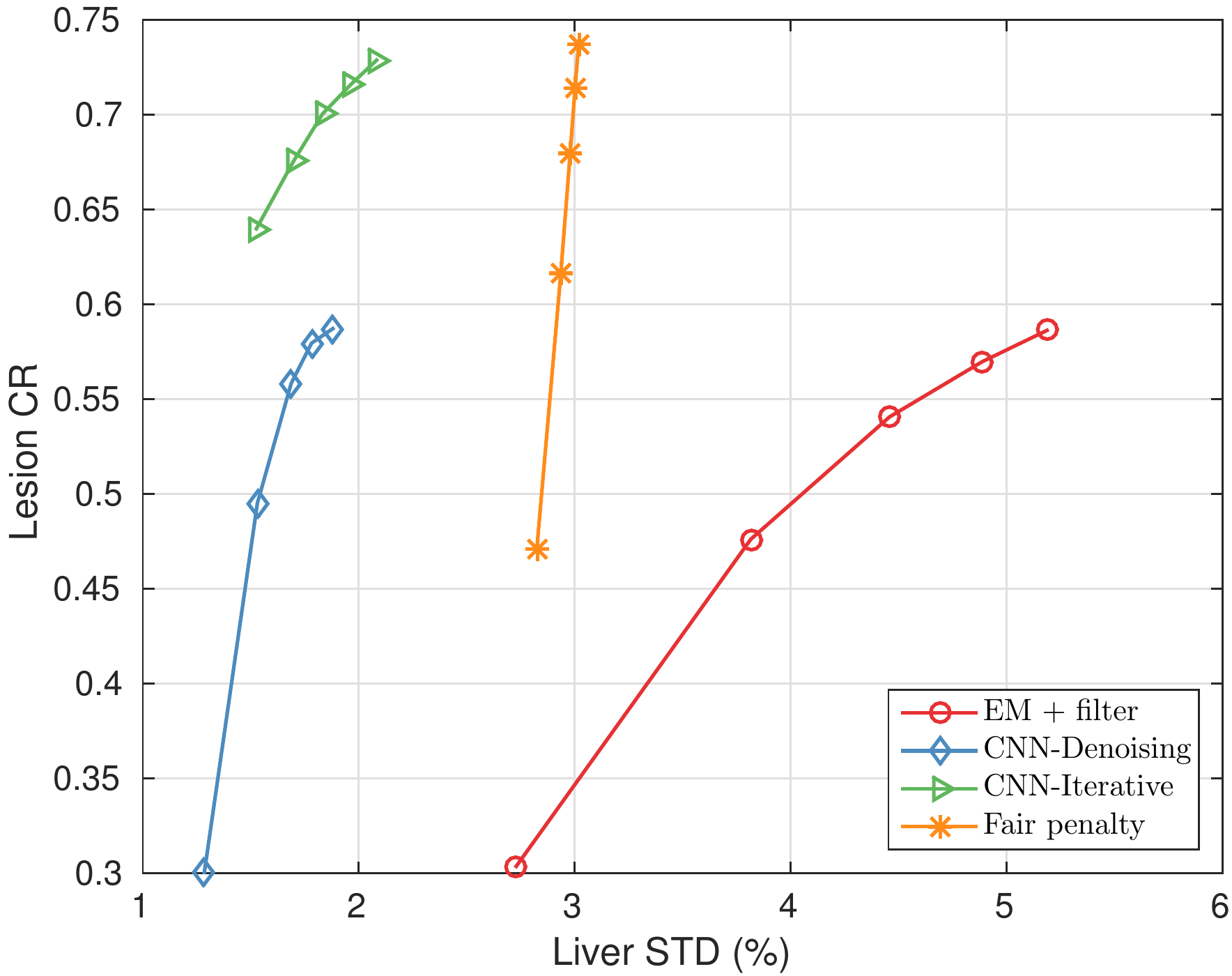}}
\caption{The contrast recovery vs. STD curves using different methods for the real data sets. }
\label{fig:real-crc-std}
\end{figure}
Fig.~\ref{fig:real-image} shows three orthogonal slices of the reconstructed images using the real data-set by different methods. We can see that the uptake of the inserted lesion in the iterative CNN method is higher than the CNN denoising method, same conclusion as in the simulation study. In addition, the iterative CNN method produced the clearest image details in the spinal regions compared with all other methods. The penalized reconstruction can preserve lesion uptake and reduce image noise, but can also present cartoon-like patterns, especially in the high uptake regions. The results using CNN methods are more natural with no obvious artifacts. The quantitative results are presented in Fig.~\ref{fig:real-crc-std}. From the figure, we can see that about two-fold STD reduction can be achieved by the CNN methods, compared with the Gaussian filtering method.
\section{Discussion}

Many prior works have used CNN in CT or MRI denoising. Here we use CNN as the image representation and embedded it into PET iterative reconstruction framework, where no prior arts exist. Compared with the CNN denoising approach, the proposed iterative CNN method has a constraint from the measured data, which can help recover some small features that are removed or annihilated by the image denoising methods. Higher contrast recovery of the lesions shown in both simulation and real data sets demonstrate this benefit.  
 
Previously the kernel method has been applied successfully in both static and dynamic PET image reconstructions. When using the kernel method, we need to explicitly specify the basis function when constructing the kernel matrix. This is not needed for CNN and the whole network representation is more data-driven. The biggest advantage of the proposed method is that more generalized prior information, such as the inter-patient scanning information, can be included in the image representation. In addition, when the prior information is from multiple resources, such as both the temporal and anatomical information, it is hard to specify how to combine those information in the kernel method.  For neural network, we can use multiple input channels to aggregate the information and let the network decide the optimum combination in the training phase. 
%
% As the network representation is a non-linear function, it can have better representation capability compared with the kernel method which uses a linear representation. Besides,

As for the optimization process of the proposed iterative CNN method, the most challenging part is Subproblem (\ref{eq:sub2}) as it is a non-linear problem. As the computation of the Jacobian matrix is difficult due to the platform limitation, currently we choose a first-order method with Nesterov momentum to solve it. However, it is easy to get trapped in local minimums. In our experiment, we found that if the initial value of $\bb{\alpha}$ is a uniform image, the result is very poor. In our proposed solution, we used the EM results after 30 iterations as the input, which can make the results more stable. Better optimization methods and more effective initial choosing strategies need further investigations. 

The network structure used in this study is the modified U-net structure, which is a fully convolutional network. One drawback of CNN is that it will remove some of the small structures in the final output. Though our proposed iterative framework can overcome this issue, better network structures, which can preserve more features, can make our proposed iterative framework work better. For example, our proposed approach can be also fit for GAN. After the generator network is trained through GAN, it can be included into the iterative framework based on the proposed method. Besides, though the data model used here is PET, it can also be used in CT or MRI reconstruction framework.

%Based on our observation, using the residual network can better preserve the image structures. But we also noticed higher noise in the final output as compared with not using residual network. As the PET data follows the Poisson model, the noise showing up in the reconstructed image is more intensity related.  Hence residual learning will be less effective as compared with CT denoising. 

\section{Conclusion}
In this work, we proposed an iterative PET image reconstruction framework by using convolutional neural network representation. Both simulated XCAT data and real data sets were used in the evaluation. Quantitative results show that the proposed iterative CNN method performs better than the CNN denoising method as well as the Gaussian filter and penalized reconstruction methods regarding contrast recovery vs. noise trade-offs. Future work will focus on more clinical data sets evaluation.

\end{twocolumn}
\bibliographystyle{IEEEtran}
\balance
\bibliography{resolution_note}

\end{document}